\documentclass[11pt]{article}

\usepackage[preprint]{acl}

\usepackage{times}
\usepackage{latexsym}

\usepackage[T1]{fontenc}
\usepackage[utf8]{inputenc}

\usepackage{microtype}

\usepackage{inconsolata}

\usepackage{graphicx}

\usepackage{amsmath}

\usepackage[utf8]{inputenc}
\usepackage{booktabs}
\usepackage{tabularx}
\usepackage{array}
\usepackage{microtype}
\usepackage{geometry}

\usepackage{float}

\usepackage{hyperref}
\usepackage{threeparttable} 

\usepackage{xeCJK} % for Chinese/Japanese/Korean characters

\usepackage{xcolor}
\usepackage{tcolorbox}
\tcbuselibrary{skins, breakable}
\definecolor{myblue}{RGB}{0, 105, 180}
\definecolor{myred}{RGB}{200, 50, 50}
\definecolor{mygray}{RGB}{240, 240, 240}

\title{Prompt Minimization: Reducing Input Redundancy Without Sacrificing Output Fidelity}

\author{
 \textbf{Marius~F.~R.~Juston\textsuperscript{1}},
 \textbf{Kevin~A.~Karim\textsuperscript{1,2}},
 \textbf{Jonathan~Gao\textsuperscript{1}},
 \textbf{Kevin~C.~Li\textsuperscript{1}},
\\
 \textbf{Rudhi~Bashambu\textsuperscript{1}}
\\
\\
 \textsuperscript{1}UIUC University of Illinois Urbana-Champaign, Illinois, USA
 \\
 \textsuperscript{2}KTH Royal Institute of Technology, Stockholm, Sweden
\\
 \small{
   \{mjuston2, kakarim2, jg48, kcli2, rudhib2\}@illinois.edu
    }
}

\begin{document}
\maketitle

\begin{abstract}
% Due on December 12\\
% Submissions should include code, (possibly) demonstrations, and the
% report.\\
% Report in paper format, should answer important questions:\\
% - What is the problem, and why is it an important problem?\\
% - What were the earlier methods, and what are their limitations?\\
% - What is new in your approach, and why did you take this approach?\\
% - What are the experimental framework decisions (datasets, metrics, implementation details) and results?\\
% - Conclusions\\
% - How did each team member contribute to the project?
Despite the growing capabilities of large language models (LLMs), prompt design remains largely heuristic and ad hoc. This project will explore \textit{prompt minimization}, the process of reducing prompts to their smallest, most information-dense form while preserving output fidelity. Practically, shorter prompts reduce computational overhead and inference latency \cite{vaswani2017attention}, especially when large contexts, such as entire documents or codebases, are included unnecessarily. Further, longer prompts can damage LLM reasoning and accuracy \cite{levy2024tasktokensimpactinput}. Theoretically, the existence of multiple prompts yielding equivalent outputs suggests a high degree of redundancy in the input space, raising fundamental questions about what information is essential to elicit specific model behaviors. We propose three variant frameworks to identify and evaluate minimal prompts and demonstrate that minimal prompts often produce outputs comparable to those of their longer counterparts. These findings suggest new directions for efficient prompt engineering and deepen our understanding of input compression in LLMs.
\end{abstract}
\section{Introduction}
% What is the problem, and why is it an important problem?

Prompting for large language models (LLMs) is often redundant: many different prompts can elicit the same or near-identical outputs, yet practitioners routinely paste long documents or sprawling instructions “just in case.” These long inputs introduce additional latency and computational overhead \cite{vaswani2017attention}, potentially resulting in substantial performance penalties for both response time and throughput. This contributes to the LLM speed and size bottlenecks in deployment, especially at scale, limiting the practical use of LLMs.\footnote{Authors have equal contribution, order determined by ChatGPT}

Further, long inputs can have damaging impacts on reasoning and accuracy. Across multiple models, lengthier prompts degrade accuracy, even when the longer prompt is just a duplication of the shorter prompt \cite{levy2024tasktokensimpactinput}. Longer context consistently decreases accuracy for multiple models, even if the model retrieves relevant information perfectly \cite{du2025contextlengthhurtsllm}. For instance, large numbers of whitespace or masked tokens, which minimally distract models, decrease accuracy, signifying an intrinsic weakness in LLMs to longer inputs. In practice, longer prompts are likely to include irrelevant information, which can distract LLMs and confuse them. Even with mitigation strategies such as chain-of-thought and self-consistency, noisy prompts consistently and significantly decrease performance \cite{shi2023largelanguagemodelseasily, wang2024resiliencelargelanguagemodels, wu2024easilyirrelevantinputsskew, jiang2025enhancingrobustnesslargelanguage}. These findings highlight LLMs' high susceptibility to distraction and their limited ability to distinguish relevant from irrelevant information.

Lastly, LLMs lack explainability, or the ability to explain how they arrived at their outputs \cite{doshi2017towards}. This has significant implications for trustworthiness. LLMs are prone to hallucinations and may inherit bias from their training process. Without explainability of how the model generated its output, users cannot be sure whether a model's output can be trusted \cite{10.1145/3639372}. Additionally, it is often challenging to design an ideal prompt for an LLM because users cannot see or understand how the model interprets and uses it. Better explainability would enable users to understand LLMs' capabilities and limitations and to design improved prompts for their end goals. Identifying shorter prompts that yield the same results as longer prompts may clarify how models interpret prompts and which information they rely on most, guiding prompt engineers in developing ideal prompts.

We target the core question behind this ambiguity: \textit{what is the minimal natural-language prompt that preserves a model’s output fidelity for a given task and model?} This matters for (i) \textbf{efficiency}—shorter inputs reduce latency, token cost, and context pressure; (ii) \textbf{accuracy}—isolating the truly relevant information mitigates LLMs' inherit difficulty in processing longer prompts and reduces distraction; and (iii) \textbf{scientific understanding}—characterizing equivalence classes of prompts reveals how models map input information to behavior.

\section{Related work}
% What were the earlier methods, and what are their limitations?

Previous work has shown that prompting with In Context Learning (ICL) can improve LLM performance on downstream tasks \cite{brown2020languagemodelsfewshotlearners}. Similarly, \citet{lester2021powerscaleparameterefficientprompt} showed that prompt tuning, in which the model parameters are fixed, and the prompt is adjusted, can achieve performance comparable to fine-tuning. The effectiveness of prompting has led to the emergence of prompt engineering as a field of study that focuses on using text instructions to extend the capabilities of LLMs \cite{sahoo2025systematicsurveypromptengineering}. Applications include using prompting to enable reasoning with Chain-of-Thought (CoT) variants as first introduced by \citet{wei2023chainofthoughtpromptingelicitsreasoning}.
 
 Another sub-field of prompt engineering deals with optimizing the prompt itself for a given task \cite{schulhoff2025promptreportsystematicsurvey}. Early approaches to prompt optimization focused on gradient-based methods \cite{shin2020autopromptelicitingknowledgelanguage}; however, due to a vast search space and large models, they become computationally infeasible and too costly to scale \cite{zhang2024sprigimprovinglargelanguage}. Later techniques include gradient-free meta prompting \cite{schulhoff2025promptreportsystematicsurvey} and genetic algorithms \cite{zhang2024sprigimprovinglargelanguage} to tune the prompt, both of which have shown promising results. 

% Reverse prompt engineering, a related line of work, focuses on reconstructing prompts from a language model's output. This problem is commonly referred to as the language model inversion problem or prompt stealing \cite{li2025reversepromptengineering,morris2023languagemodelinversion,sha2024promptstealingattackslarge}, since some applications involve \textit{stealing} or reverse engineering system prompts. While reverse prompt engineering seeks to reconstruct prompts post hoc, a natural question arises: can it be combined with prompt optimization to minimize the initial prompt? To our best knowledge, no prior work has directly addressed this question; thus, this project aims to bridge that gap.

Past studies on prompt minimization have split mainly into two approaches: pruning, or directly removing less relevant tokens, and summarization, changing the prompt itself while maintaining overall semantic meaning \cite{chang2024efficientpromptingmethodslarge}. Several pruning methods split prompts into sections (e.g., instruction, demonstration, question, etc.) and minimize the size of individual sections \cite{jiang2023llmlinguacompressingpromptsaccelerated, zhou2023efficientpromptingdynamicincontext}. Another pruning strategy is to consider token-wise importance and remove tokens deemed less important \cite{jiang2023llmlinguacompressingpromptsaccelerated, Jung_2024, li-etal-2023-compressing, ali2024promptsawleveragingrelationawaregraphs}. These techniques have been shown to minimize prompt length while retaining output fidelity successfully.

Summarization methods are relatively less explored. These methods generally rely on ICL to compress long text while preserving its meaning. Many of these summarization methods focus on compressing context in retrieval-augmented generation (RAG) \cite{du2025contextlengthhurtsllm, xu2023recompimprovingretrievalaugmentedlms, chen2023walkingmemorymazecontext, yoon2024compactcompressingretrieveddocuments}. There have been a few summarization methods focused on prompt (rather than context) minimization, including Style-Compress \cite{pu2024stylecompressllmbasedpromptcompression}, ATF \cite{jiang2025enhancingrobustnesslargelanguage}, and Nano-Capsulator \cite{chuang2024learningcompresspromptnatural}. Though these methods have shown success, they are limited — Style-Compress is deliberately task-specific; ATF is designed to remove noisy information rather than minimize relevant information; and Nano-Capsulator requires extensive training. In this study, we propose three task-agnostic prompt minimization techniques for no/low-training prompts. To our best knowledge, no prior work has directly addressed this question; thus, this project aims to bridge that gap.

\section{Approach}
% What is new in your approach, and why did you take this approach?

% TODO: determine which method(s) we want to put in our final report, create diagrams and write-ups for each
We propose three approaches for task-agnostic prompt minimization. These approaches follow the same general framework, with slight variations in how minimization is approached. Notably, all three of these approaches are task-agnostic and require relatively little training.

\subsection{Zero-Shot Learning} \label{sec:zsl}

\begin{figure}[ht!]
    \centering
    \includegraphics[width=1.0\linewidth]{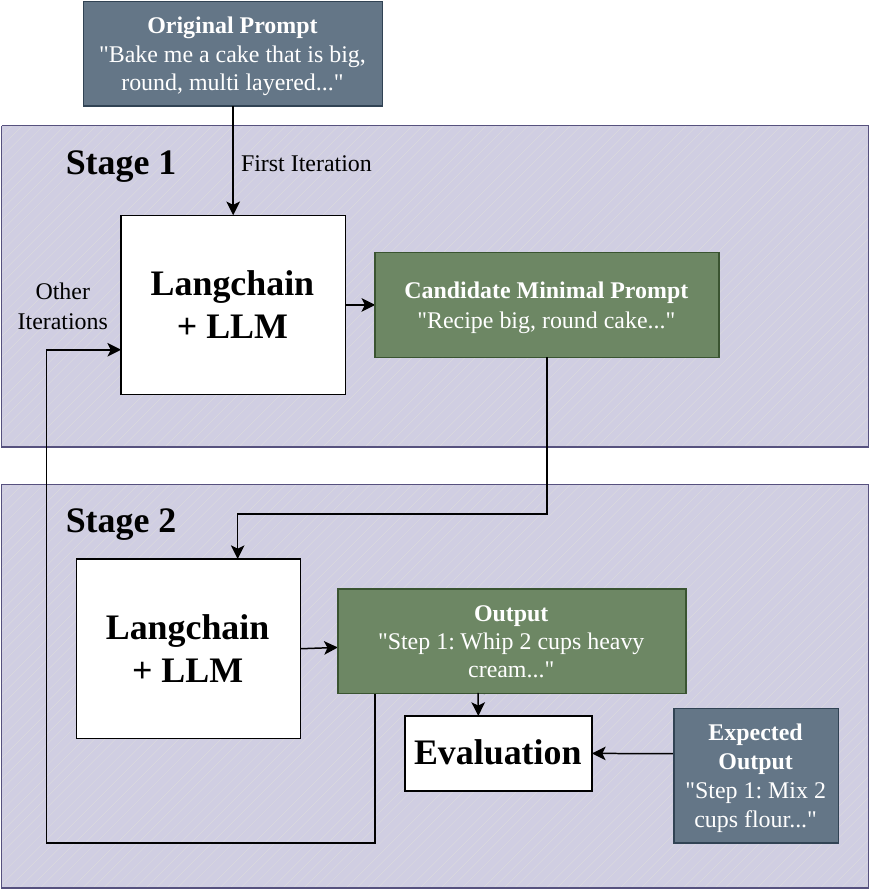}
    \caption{Flowchart illustrating proposed approach 1 (Zero-Shot Learning). We use an LLM to minimize a long prompt and then compare that minimized prompt's output to the expected output. This variant generates one candidate minimal prompt and uses zero-shot learning.}
    \label{fig:proposed_approach_1}
\end{figure}

First, as a baseline, we propose a zero-shot LLM-based minimization that consists of two steps: minimization and evaluation. See Figure \ref{fig:proposed_approach_1}. Given an Original Prompt, we ask an LLM to condense the prompt as much as possible. Then, the new prompt and its output are compared to the original prompt and output. Both stages are connected in a langchain pipeline that automates the minimization and evaluation process.  

During stage (1), the minimization phase, the LLM is fed the original prompt and instructed to find the minimal possible candidate prompt that still preserves its core meaning. This approach is considered zero-shot because the only information provided is the original prompt, and the LLM is a system prompt engineered to complete the task in a single step with no reasoning.  

After a candidate prompt is generated, stage (2) evaluates the prompt compared to the original prompt. First, the candidate prompt is presented to a chat-completion LLM, which generates a new output. Then, the candidate prompt and its corresponding output are compared to the original prompt and output. The comparison considers both the compression ratio of the prompts (by length) and the BERT similarity of the outputs. 

Using the generated candidate prompt as input to stage one, it is possible to find an even smaller representation. Because the entire process is automated, the approach can be iterated until a desired stopping criterion is met.
\subsection{In-Context Learning} \label{sec:icl}
% Our second approach follows the same general framework, but generates multiple candidate minimal prompts and outputs. We propose a task-agnostic, multi-shot prompt minimization approach. Our approach has two stages: given a prompt and its expected output, we (1) use an LLM to generate candidate minimal prompts and (2) pass these candidate minimal prompts into an LLM to evaluate their outputs on size, similarity, etc. We iteratively repeat this process to find an ideal minimal prompt. Figure \ref{fig:proposed_approach_2} illustrates this process.

\begin{figure}[ht!]
    \centering
    \includegraphics[width=1.0\linewidth]{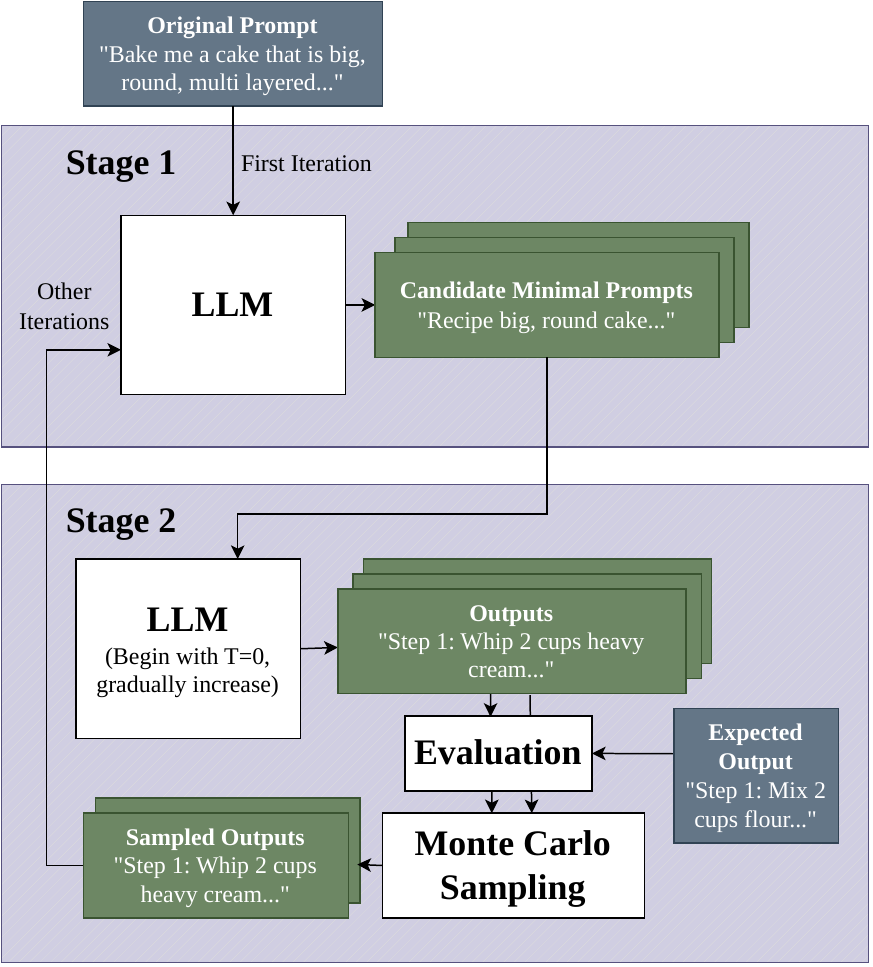}
    \caption{Flowchart illustrating proposed approach 2 (In-Context Learning). We use an LLM to minimize a long prompt, then compare the resulting prompt's output to the expected output. This variant generates multiple candidate minimal prompts and uses in-context learning.}
    \label{fig:proposed_approach_2}
\end{figure}

Our second approach extends the zero-shot framework by formulating prompt minimization as a discrete optimization problem solved via an evolutionary strategy. Unlike the zero-shot approach, which generates a single trajectory of reductions, this method maintains a frontier of candidate minimal prompts to avoid local minima. We employ an LLM using In-Context Learning (ICL) to generate diverse prompt candidates. Figure \ref{fig:proposed_approach_2} illustrates the two-stage optimization process across $T$ iterations. 

For stage (1), we generate a new generation of candidate prompts. For the initial iteration, we derive candidates directly from the original prompt. For future iterations, we seed the prompts by sampling from the historical population of top prompts. To balance exploration (diversity) and exploitation (quality), we employ a weighted sampling selection mechanism based on inverse-fitness weighting, 
\begin{equation*}
    P(p_i) \propto \frac{1}{\text{Score}(p_i) + 1},
\end{equation*}
such that a lower score implies a better minimal prompt. 

These seed parent prompts undergo a mutation process. Similar to the zero-shot approach, the LLM is fed a system prompt containing potential insights for valid compressions (e.g., removing politeness markers, condensing instructions) to guide it in compressing the selected parent prompt while retaining semantic intent.

Stage (2) evaluates the newly generated candidate prompts by passing them to the same LLM to generate the outputs. These outputs are scored in the same way as the zero-shot approach, using a weighted objective function that combines the compression ratio and BERTScore semantic similarity. We then apply truncation selection (elitism) at the end of each iteration: only the global top-$N$ prompts are retained in the priority queue to serve as potential seeds for the next iteration. This cycle continues until the maximum iteration count is reached or convergence is detected.

% Our second approach follows the same general framework, but generates multiple candidate minimal prompts and outputs using in-context learning. Figure \ref{fig:proposed_approach_2} illustrates this process.

% For stage (1), we generate a set of candidate minimal prompts from the original prompt (for the first iteration) or from a sample of prompts from the previous iteration (for all other iterations). We seed prompt generation with randomly selected prompts from the historical top prompts; the selection is based on an inverse weighting scheme: the smaller the score, the more likely it is to be selected. To keep the search space maintainable, we use Monte Carlo sampling to select the prompts that are further minimized in the next iteration. Only the historically top N best prompts are used moving forward. This approach converges towards ideal minimal prompts while exploring the sample space and maintaining efficiency.

% Genetic algorithms inspire our approach for stage (2) to balance efficiency and robustness. We first generate the outputs from newly generated prompt candidates, and then score and sort them. The sorted prompts then get fed back into stage (1). Iteration continues until further prompt minimization degrades output fidelity or a stopping criterion is met.
%
%
% This approach differs from prior work by focusing on minimizing prompts themselves, with no constraints on their nature. Our approach requires no training and works flexibly on different tasks and problem domains.

% \begin{figure*}
%     \centering
%     \includegraphics[width=1\linewidth]{Figs/Project_Proposal-prompt examples.drawio.png}
%     \caption{Example prompts and outputs over iterations, emphases added.}
%     \label{fig:placeholder}
% \end{figure*}

\subsection{RL-Fine Tuning} \label{sec:rl}
\begin{figure}[ht!]
    \centering
    \includegraphics[width=1.0\linewidth]{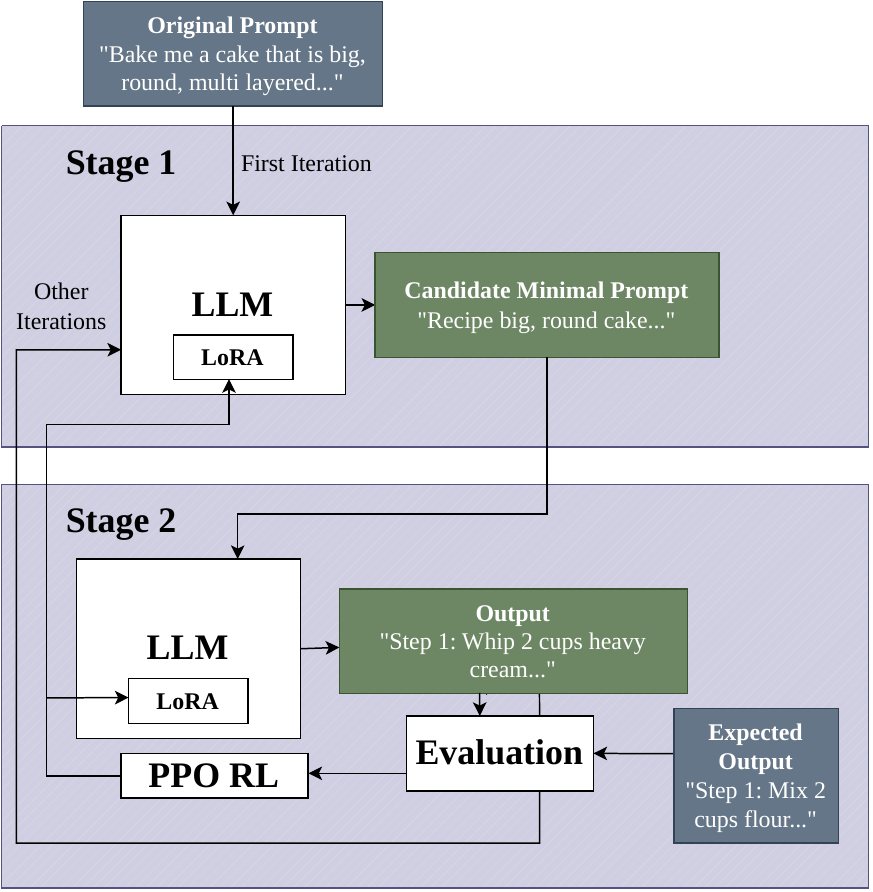}
    \caption{Flowchart illustrating proposed approach 3 (RL-Fine Tuning). We use an LLM to minimize a long prompt and then compare that minimized prompt's output to the expected output. This variant fine-tunes the LLM using LoRA and PPO RL.}
    \label{fig:proposed_approach_3}
\end{figure}

Lastly, we propose a reinforcement learning (RL)- based framework to train an automated prompt compressor that learns a robust rewriting policy. This approach uses Proximal Policy Optimization (PPO) to fine-tune an LLM for semantic compression. To maintain computational efficiency and preserve the model's general reasoning and language capabilities, we freeze the pre-trained backbone and optimize only a set of low-rank weights (LoRA). Figure \ref{fig:proposed_approach_3} illustrates this training pipeline. \newline
Our training process consists of two stages: (1) Policy Rollout and (2) Reward Calculation and Optimization. 

In stage (1), the model, acting as the policy, generates a compressed version of the prompt. Unlike standard decoding, we inject dynamic systems instructions that evolve based on previous performance. For example, if the policy fails to compress or loses semantic meaning in a prior step sufficiently, explicit constraints are added to the context window. This mimics a ``chain of thought" correction process, guiding the exploration through the solution space. 

In stage (2), we evaluate the generated candidate using a composite reward function that balances semantic fidelity and token reduction. We use BERTScore to measure the embedding alignment between the original and generated outputs, ensuring instruction adherence while simultaneously calculating a compression ratio score. These signals are compressed into a scalar reward, which is used to compute the advantage and to update the LoRA adapters via PPO. This allows the model to gradually converge on a policy that aggressively removes redundancy while strictly maintaining the original prompt's intent. 
\section{Experiments}
% What are the experimental framework decisions (datasets, metrics, implementation details) and results?

\subsection{Dataset Curation}
Due to the limited availability of long, open-ended natural-language question prompts, we used OpenAI's GPT-4o and GPT-5.1, Grok, and Claude AI's Sonnet 4.6 to generate 60 long, open-ended prompts. The prompts ranged from 60 to 232 words, with an average of 128. These prompts span a wide range of domains, such as mathematical problem-solving, interpersonal relationship challenges, creative writing, and more.

The system prompt used to generate the dataset was: 
\begin{tcolorbox}[
    colback=mygray, 
    colframe=gray!50!black, 
    title=\textbf{Data Generation Prompt}, 
    fonttitle=\sffamily\small,
    boxrule=0.5pt,
    left=4pt, right=4pt, top=4pt, bottom=4pt
]
\small\sffamily\textit{Can you generate 10 example prompts that I could use for a prompt-minimization dataset? The ideal prompt to minimize would be an open-ended question with context, so that it could be minimized. Make the prompts span a wide range of topics, such as math, law, engineering, creative writing, finance, and more. The prompt should be at least 250 words. Provide just the initial prompt.
}
\end{tcolorbox}
The full dataset of prompts is available for viewing and downloading in the code repository.

\subsection{Evaluation Metrics}
Our evaluation focuses on two key factors: semantic similarity and length compression.

We denote $\mathbf{y}_0$ as the initial prompt's generated output and $\mathbf{y}_i$ as the generated output for the $i$th prompt.
\subsubsection{Semantic similarity}
We used BERTScore \cite{zhang2020bertscoreevaluatingtextgeneration} to measure semantic similarity. 

% TODO: finish this description

 In our experiment, we use this to evaluate whether the generated outputs hold the same meaning, regardless of specific sentence structure or word choice.
    
\subsubsection{Compression}

We determine length compression as the ratio of the lengths of the compressed prompt and the original prompt, or:
$$\text{Compression Score}(\textbf{y}_i) =\frac{|\mathbf{y}_i|}{|\mathbf{y}_0|}$$
where $|\cdot|$ denotes the length (in characters) of a string.

In our experiment, we use this to evaluate the compression level of a candidate minimal prompt.
\begin{table*}[htb]
\centering
\begin{threeparttable}
\footnotesize
\begin{tabular}{@{}lcccccc@{}}
\toprule
 & \multicolumn{2}{c}{Llama 3.1 8B} & \multicolumn{2}{c}{Qwen2.5 32B} \\ \cmidrule(lr){2-3} \cmidrule(lr){4-5}
Algorithm & Comp & BERT & Comp & BERT \\
\midrule
Zero-Shot Learning & 0.08 $\pm$ 0.05 & 0.88 $\pm$ 0.03 &\textbf{ 0.17 $\pm$ 0.08} & 0.89 $\pm$ 0.02 \\ 
In-Context Learning & \textbf{0.04 $\pm$ 0.02} & 0.89 $\pm$ 0.02 & 0.19 $\pm$ 0.10 & 0.90 $\pm$ 0.01 \\
RL Fine-Tuning & 0.22 $\pm$ 0.11 & 0.89 $\pm$ 0.02 & 0.25 $\pm$ 0.09 & 0.89 $\pm$ 0.02 \\
\bottomrule
\end{tabular}
\end{threeparttable}
\caption[Mean Accuracy]{Aggregate statistics for the different algorithms using the different LLMs}
\label{tab:agg_stats}
\end{table*}

\section{Conclusions}
This work studied prompt minimization: finding short, information-dense natural language prompts that preserve a model’s output behavior. We presented three complementary procedures—zero-shot search, in-context evolutionary search, and an RL-flavored variant — and evaluated them using a two-term objective that trades off semantic fidelity and compression. Across diverse, long-form instructions, the procedures reliably discovered substantially shorter prompts whose outputs remained close in meaning to those of the originals. With the baseline explicitly set to iteration~0 and a baseline score of 0.5, we observe a sharp improvement in the next one to three iterations, followed by diminishing returns; the trajectory visualizations make these phases evident (Figs.~\ref{fig:traj_llama},~\ref{fig:traj_qwen1}). Taken together, our results suggest that a significant fraction of typical prompts is redundant and that simple, training-light search already recovers strong compressions.

\subsection{Takeaways}
\begin{figure}[ht!]
    \centering
    \includegraphics[width=1\linewidth]{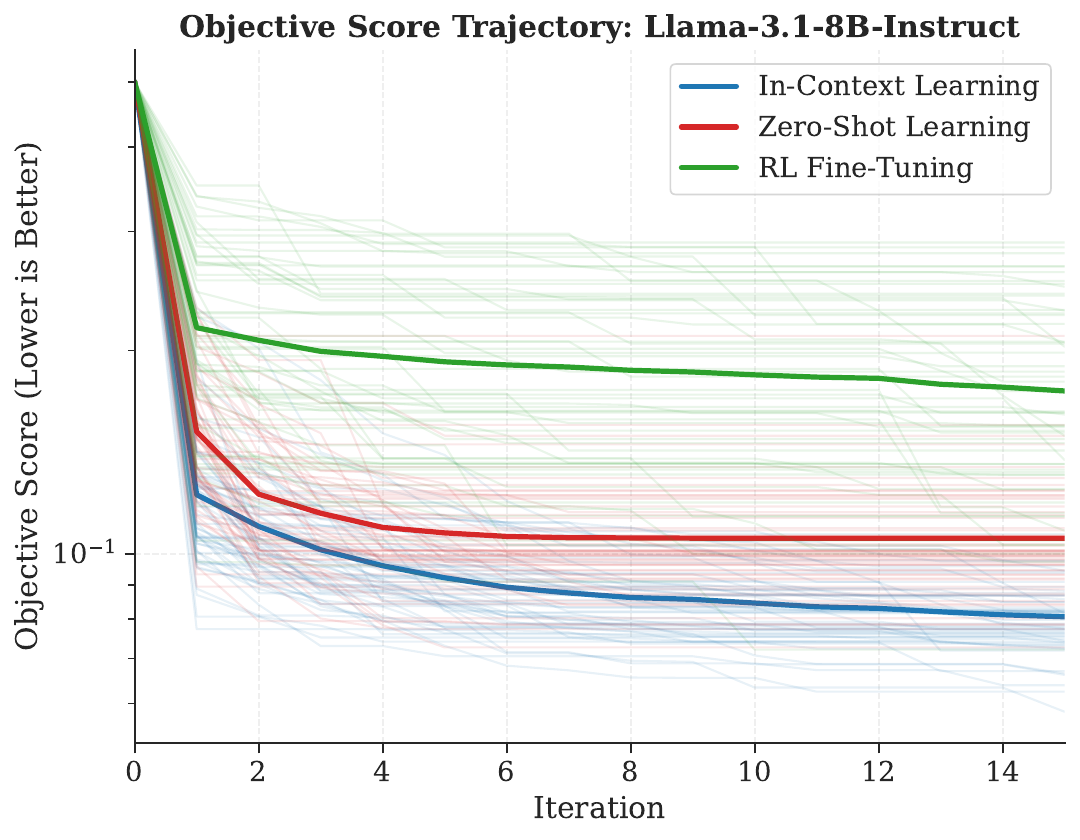}
    \caption{Best score trajectory of the 60 long context prompts using the meta-llama/Llama-3.1-8B-Instruct LLM. BERTScore and compression score are weighted equally at 0.5.}
    \label{fig:traj_llama}
\end{figure}

\begin{figure}[ht!]
    \centering
    \includegraphics[width=1\linewidth]{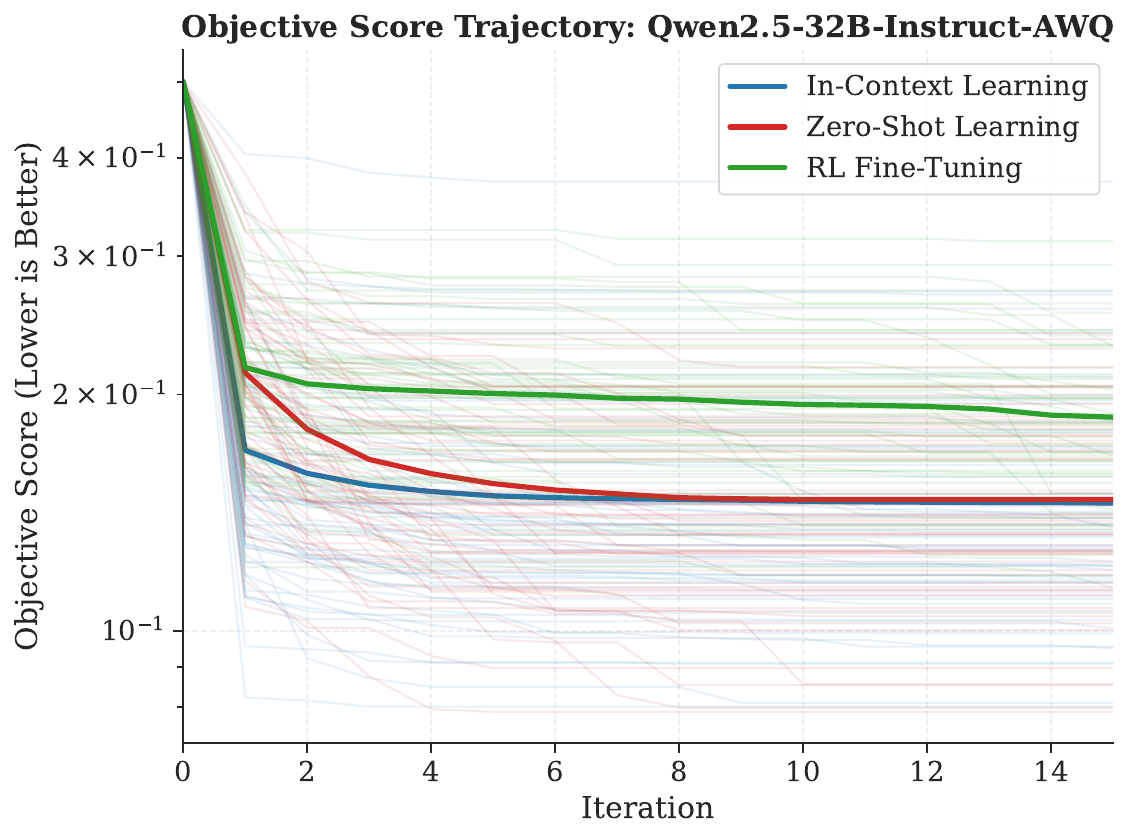}
    \caption{Best score trajectory of the 60 long context prompts using the Qwen/Qwen2.5-32B-Instruct-AWQ LLM. BERTScore and compression score are weighted equally at 0.5.}
    \label{fig:traj_qwen1}
\end{figure}

\begin{figure}[ht!]
    \centering
    \includegraphics[width=1\linewidth]{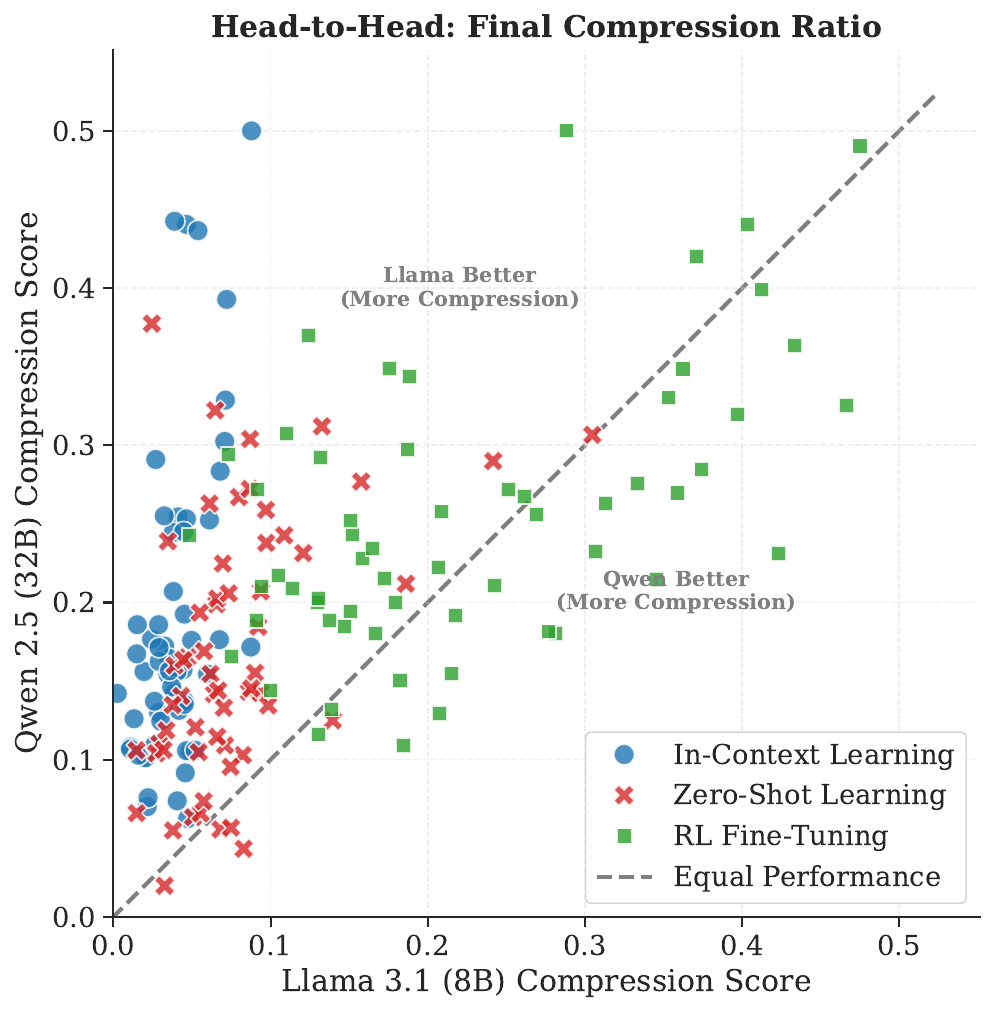}
    \caption{Best compression score compared between Qwen/Qwen2.5-32B-Instruct-AWQ and meta-llama/Llama-3.1-8B-Instruct for the same prompt. Lower is better.}
    \label{fig:traj_qwen2}
\end{figure}

\begin{figure}[ht!]
    \centering
    \includegraphics[width=1\linewidth]{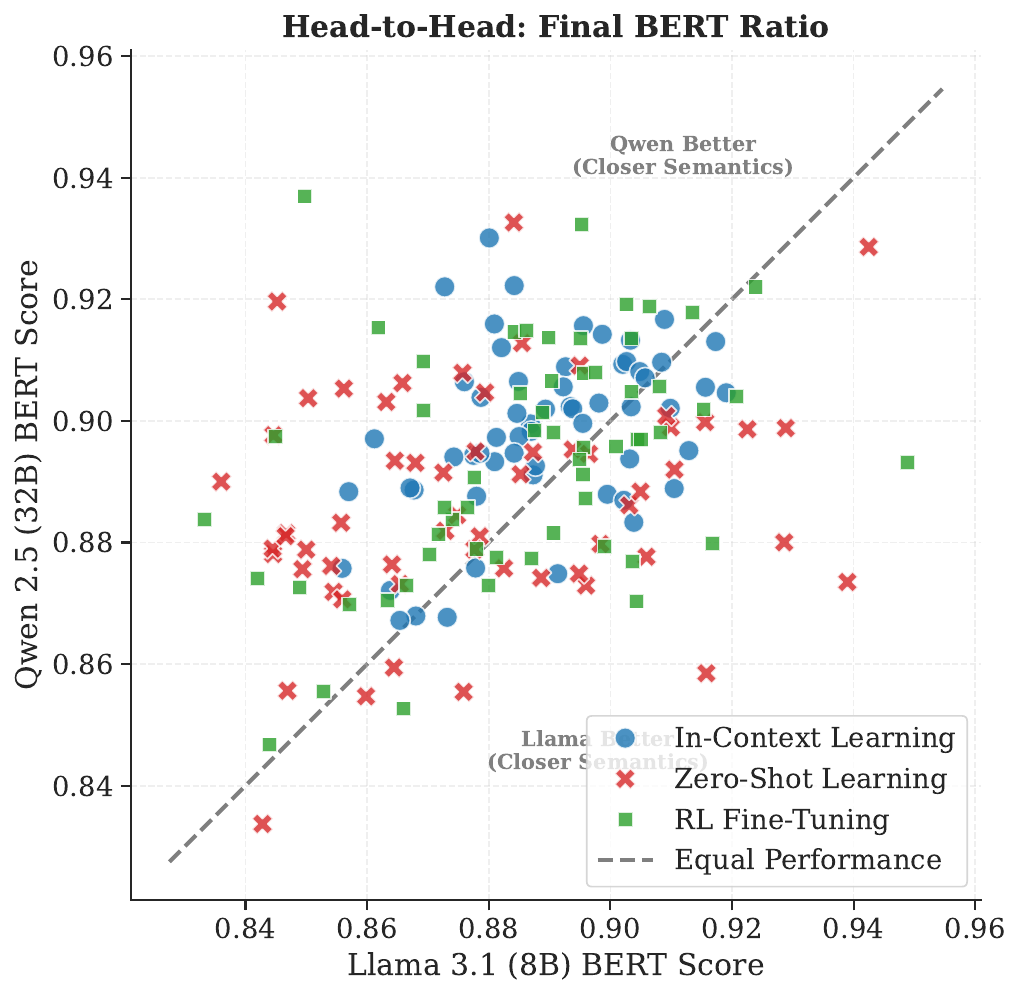}
    \caption{Best BERT score compared between Qwen/Qwen2.5-32B-Instruct-AWQ and meta-llama/Llama-3.1-8B-Instruct for the same prompt. Lower is better.}
    \label{fig:traj_qwen3}
\end{figure}

Two themes recur across models and methods. First, most gains arrive early: after the baseline (iteration~0), total loss typically drops steeply within a few steps before plateauing, as reflected in Figs.~\ref{fig:traj_llama} and \ref{fig:traj_qwen1}. 
Second, compression behavior is model-dependent: under matched settings, Llama-3.1-8B achieves stronger compression than Qwen-2.5-32B at a similar BERTScore (Table~\ref{tab:agg_stats}). This suggests that smaller models may compress more aggressively while still preserving semantic similarity. Methodologically, keeping a small frontier and resampling around it produced more stable progress than pure zero-shot proposals. Finally, the fixed 0.5/0.5 weighting between compression and similarity clearly shaped the discovered frontier: heavier emphasis on similarity preserves structure, whereas heavier emphasis on compression would push toward terser prompts at the risk of meaning drift. Moreover, the cross-model scatter (as reflected in Figs. ~\ref{fig:traj_qwen2} and ~\ref{fig:traj_qwen3}) shows Llama-3.1-8B reaches lower compression ratios while keeping BERTScore effectively on par with Qwen-2.5-32B, i.e., stronger compression at comparable semantics.
\subsection{Limitations}
Iterative search with multiple candidates per step can be expensive in wall-clock time and tokens, especially when temperatures or frontier sizes are swept. Our fidelity estimates rely on embedding/BERT-style similarity, which may miss pragmatic nuances and can over- or under-penalize paraphrases. Several of the long prompts were machine-generated; outcomes may differ for expert-authored prompts with strict correctness constraints. Finally, a minimal prompt that reproduces a single reference output may not generalize to small task perturbations or broader input distributions.

\subsection{Future Work}
A natural next step is to complement automatic similarity with human preference studies or lightweight preference models and to add an explicit quality/helpfulness term to the objective. Rather than a single weighted sum, we aim to expose a Pareto frontier over similarity, compression, cost, and safety so practitioners can choose operating points. Richer fidelity signals, such as NLI consistency, factuality checks, or task-specific scorers, could reduce metric gaming. Cost can be reduced with adaptive early stopping and per-prompt candidate budgets learned from online improvement rates. Finally, conditioning the minimizer on task type and model family, extending minimization to retrieved contexts and system prompts with safety filters, and packaging the visualizer and scripts into a small library with a short practitioner handout should make these techniques easy to adopt in practice.

\subsection{Code availability}

The code is available at the GitHub repo \href{https://github.com/jontgao/546-prompt-minimization}{jontgao/546-prompt-minimization}

The output logs and runs are available at the HuggingFace reposity \href{https://huggingface.co/SuperComputer/PromptMinimization}{SuperComputer/PromptMinimization}

\bibliography{custom}

\appendix

\section{Result examples}

\noindent \textbf{Zero-Shot Learning Examples:} Figures \ref{fig:milestones_11334092_karim} and \ref{fig:milestones_4877789_karim} use the Algorithm in \ref{sec:zsl} for generating the minimization of the prompts.

\begin{figure*}[t]
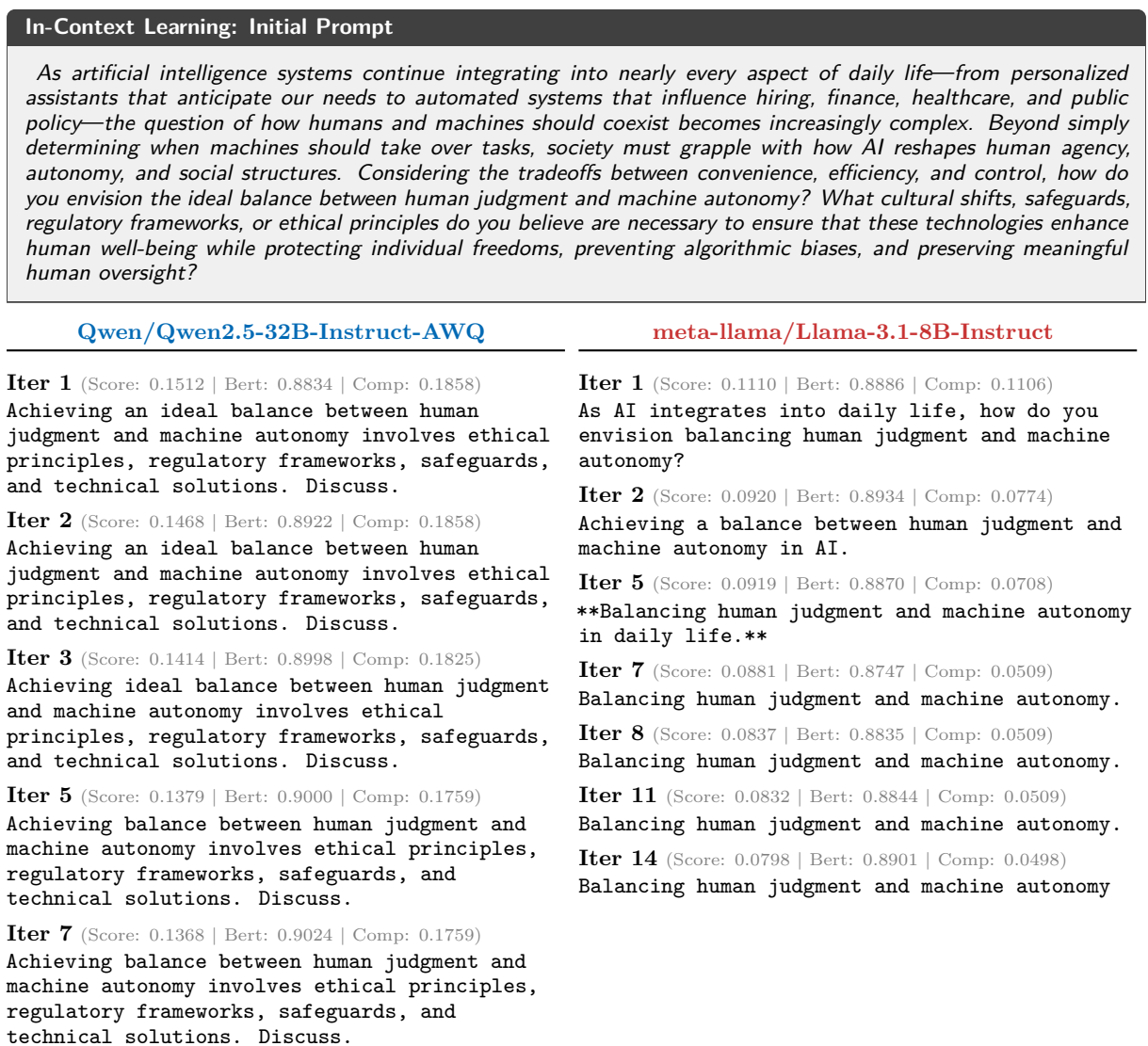

\centering
\begin{tcolorbox}[
    colback=mygray, 
    colframe=gray!50!black, 
    title=\textbf{In-Context Learning: Initial Prompt}, 
    fonttitle=\sffamily\small, 
    sharp corners=south,
    boxrule=0.5pt,
    left=4pt, right=4pt, top=4pt, bottom=4pt
]
\small\sffamily\textit{ As artificial intelligence systems continue integrating into nearly every aspect of daily life—from personalized assistants that anticipate our needs to automated systems that influence hiring, finance, healthcare, and public policy—the question of how humans and machines should coexist becomes increasingly complex. Beyond simply determining when machines should take over tasks, society must grapple with how AI reshapes human agency, autonomy, and social structures. Considering the tradeoffs between convenience, efficiency, and control, how do you envision the ideal balance between human judgment and machine autonomy? What cultural shifts, safeguards, regulatory frameworks, or ethical principles do you believe are necessary to ensure that these technologies enhance human well-being while protecting individual freedoms, preventing algorithmic biases, and preserving meaningful human oversight?
}
\end{tcolorbox}
\vspace{-2pt}
\begin{minipage}[t]{0.49\textwidth}
\centering
\textcolor{myblue}{\textbf{\small Qwen/Qwen2.5-32B-Instruct-AWQ}}
\vspace{2pt}
\hrule height 0.8pt
\vspace{4pt}
\begin{flushleft}
\footnotesize
\textbf{Iter 1} {\scriptsize \color{gray} (Score: 0.1512 $\mid$ Bert: 0.8834 $\mid$ Comp: 0.1858)} \\ 
\vspace{1pt}
\texttt{Achieving an ideal balance between human judgment and machine autonomy involves ethical principles, regulatory frameworks, safeguards, and technical solutions. Discuss.}\\
\vspace{4pt}
\textbf{Iter 2} {\scriptsize \color{gray} (Score: 0.1468 $\mid$ Bert: 0.8922 $\mid$ Comp: 0.1858)} \\ 
\vspace{1pt}
\texttt{Achieving an ideal balance between human judgment and machine autonomy involves ethical principles, regulatory frameworks, safeguards, and technical solutions. Discuss.}\\
\vspace{4pt}
\textbf{Iter 3} {\scriptsize \color{gray} (Score: 0.1414 $\mid$ Bert: 0.8998 $\mid$ Comp: 0.1825)} \\ 
\vspace{1pt}
\texttt{Achieving ideal balance between human judgment and machine autonomy involves ethical principles, regulatory frameworks, safeguards, and technical solutions. Discuss.}\\
\vspace{4pt}
\textbf{Iter 5} {\scriptsize \color{gray} (Score: 0.1379 $\mid$ Bert: 0.9000 $\mid$ Comp: 0.1759)} \\ 
\vspace{1pt}
\texttt{Achieving balance between human judgment and machine autonomy involves ethical principles, regulatory frameworks, safeguards, and technical solutions. Discuss.}\\
\vspace{4pt}
\textbf{Iter 7} {\scriptsize \color{gray} (Score: 0.1368 $\mid$ Bert: 0.9024 $\mid$ Comp: 0.1759)} \\ 
\vspace{1pt}
\texttt{Achieving balance between human judgment and machine autonomy involves ethical principles, regulatory frameworks, safeguards, and technical solutions. Discuss.}
\end{flushleft}
\end{minipage}
\hfill
\begin{minipage}[t]{0.49\textwidth}
\centering
\textcolor{myred}{\textbf{\small meta-llama/Llama-3.1-8B-Instruct}}
\vspace{2pt}
\hrule height 0.8pt
\vspace{4pt}
\begin{flushleft}
\footnotesize
\textbf{Iter 1} {\scriptsize \color{gray} (Score: 0.1110 $\mid$ Bert: 0.8886 $\mid$ Comp: 0.1106)} \\ 
\vspace{1pt}
\texttt{As AI integrates into daily life, how do you envision balancing human judgment and machine autonomy?}\\
\vspace{4pt}
\textbf{Iter 2} {\scriptsize \color{gray} (Score: 0.0920 $\mid$ Bert: 0.8934 $\mid$ Comp: 0.0774)} \\ 
\vspace{1pt}
\texttt{Achieving a balance between human judgment and machine autonomy in AI.}\\
\vspace{4pt}
\textbf{Iter 5} {\scriptsize \color{gray} (Score: 0.0919 $\mid$ Bert: 0.8870 $\mid$ Comp: 0.0708)} \\ 
\vspace{1pt}
\texttt{**Balancing human judgment and machine autonomy in daily life.**}\\
\vspace{4pt}
\textbf{Iter 7} {\scriptsize \color{gray} (Score: 0.0881 $\mid$ Bert: 0.8747 $\mid$ Comp: 0.0509)} \\ 
\vspace{1pt}
\texttt{Balancing human judgment and machine autonomy.}\\
\vspace{4pt}
\textbf{Iter 8} {\scriptsize \color{gray} (Score: 0.0837 $\mid$ Bert: 0.8835 $\mid$ Comp: 0.0509)} \\ 
\vspace{1pt}
\texttt{Balancing human judgment and machine autonomy.}\\
\vspace{4pt}
\textbf{Iter 11} {\scriptsize \color{gray} (Score: 0.0832 $\mid$ Bert: 0.8844 $\mid$ Comp: 0.0509)} \\ 
\vspace{1pt}
\texttt{Balancing human judgment and machine autonomy.}\\
\vspace{4pt}
\textbf{Iter 14} {\scriptsize \color{gray} (Score: 0.0798 $\mid$ Bert: 0.8901 $\mid$ Comp: 0.0498)} \\ 
\vspace{1pt}
\texttt{Balancing human judgment and machine autonomy}
\end{flushleft}
\end{minipage}
\vspace{6pt}
\caption{\textbf{Milestone Discoveries in Prompt Minimization using In-Context Learning.} The top panel shows the verbose initial prompt. The bottom panels compare the minimization trajectory of Qwen-32B (Left) and Llama-3.1-8B (Right) using Algorithm from Section \ref{sec:icl}. Metrics indicate total score, BERT-score, and compression ratio. Running with Temperature 0.9.}
\label{fig:milestones_11334092_marius}
\end{figure*}

\noindent \textbf{In-Context Learning Examples:} Figures \ref{fig:milestones_11334092_marius} and \ref{fig:milestones_4877789_marius} use the Algorithm in \ref{sec:icl} for generating the minimization of the prompts. 

\noindent \textbf{RL Fine-Tuning Examples:} Figures \ref{fig:milestones_5129854_li} and \ref{fig:milestones_59082115_li} use the Algorithm in \ref{sec:rl} for generating the minimization of the prompts. 

\begin{figure*}[t]
\centering
\begin{tcolorbox}[
    colback=mygray, 
    colframe=gray!50!black, 
    title=\textbf{Zero-Shot Learning: Initial Prompt}, 
    fonttitle=\sffamily\small, 
    sharp corners=south,
    boxrule=0.5pt,
    left=4pt, right=4pt, top=4pt, bottom=4pt
]
\small\sffamily\textit{ As artificial intelligence systems continue integrating into nearly every aspect of daily life—from personalized assistants that anticipate our needs to automated systems that influence hiring, finance, healthcare, and public policy—the question of how humans and machines should coexist becomes increasingly complex. Beyond simply determining when machines should take over tasks, society must grapple with how AI reshapes human agency, autonomy, and social structures. Considering the tradeoffs between convenience, efficiency, and control, how do you envision the ideal balance between human judgment and machine autonomy? What cultural shifts, safeguards, regulatory frameworks, or ethical principles do you believe are necessary to ensure that these technologies enhance human well-being while protecting individual freedoms, preventing algorithmic biases, and preserving meaningful human oversight?
}
\end{tcolorbox}
\vspace{-2pt}
\begin{minipage}[t]{0.49\textwidth}
\centering
\textcolor{myblue}{\textbf{\small Qwen/Qwen2.5-32B-Instruct-AWQ}}
\vspace{2pt}
\hrule height 0.8pt
\vspace{4pt}
\begin{flushleft}
\footnotesize
\textbf{Iter 1} {\scriptsize \color{gray} (Score: 0.1648 $\mid$ Bert: 0.8806 $\mid$ Comp: 0.2102)} \\ 
\vspace{1pt}
\texttt{How can we balance human judgment and AI autonomy to enhance well-being while preventing biases and ensuring oversight? Discuss necessary cultural shifts, safeguards, and ethical principles.}\\
\vspace{4pt}
\textbf{Iter 2} {\scriptsize \color{gray} (Score: 0.1337 $\mid$ Bert: 0.8864 $\mid$ Comp: 0.1538)} \\ 
\vspace{1pt}
\texttt{Balance human judgment and AI to enhance well-being, prevent biases, and ensure oversight. Discuss cultural shifts, safeguards, and ethics.}\\
\vspace{4pt}
\textbf{Iter 3} {\scriptsize \color{gray} (Score: 0.1318 $\mid$ Bert: 0.8912 $\mid$ Comp: 0.1549)} \\ 
\vspace{1pt}
\texttt{Balance human judgment with AI to enhance well-being, prevent biases, and ensure oversight. Discuss cultural shifts, safeguards, and ethics.}\\
\vspace{4pt}
\textbf{Iter 4} {\scriptsize \color{gray} (Score: 0.1257 $\mid$ Bert: 0.8892 $\mid$ Comp: 0.1405)} \\ 
\vspace{1pt}
\texttt{Balance human judgment with AI for well-being, bias prevention, and oversight. Discuss cultural shifts, safeguards, and ethics.}
\end{flushleft}
\end{minipage}
\hfill
\begin{minipage}[t]{0.49\textwidth}
\centering
\textcolor{myred}{\textbf{\small meta-llama/Llama-3.1-8B-Instruct}}
\vspace{2pt}
\hrule height 0.8pt
\vspace{4pt}
\begin{flushleft}
\footnotesize
\textbf{Iter 1} {\scriptsize \color{gray} (Score: 0.1299 $\mid$ Bert: 0.8851 $\mid$ Comp: 0.1449)} \\ 
\vspace{1pt}
\texttt{How can humans and AI systems balance autonomy and agency in a way that enhances human well-being and protects individual freedoms?}\\
\vspace{4pt}
\textbf{Iter 2} {\scriptsize \color{gray} (Score: 0.1055 $\mid$ Bert: 0.8852 $\mid$ Comp: 0.0962)} \\ 
\vspace{1pt}
\texttt{Balancing AI autonomy and human agency for enhanced well-being and individual freedoms.}\\
\vspace{4pt}
\textbf{Iter 3} {\scriptsize \color{gray} (Score: 0.0943 $\mid$ Bert: 0.8844 $\mid$ Comp: 0.0730)} \\ 
\vspace{1pt}
\texttt{Balancing AI autonomy and human agency for well-being and freedom.}\\
\vspace{4pt}
\textbf{Iter 4} {\scriptsize \color{gray} (Score: 0.0803 $\mid$ Bert: 0.8826 $\mid$ Comp: 0.0431)} \\ 
\vspace{1pt}
\texttt{Balancing AI autonomy and human agency.}
\end{flushleft}
\end{minipage}
\vspace{6pt}
\caption{\textbf{Milestone Discoveries in Prompt Minimization using Zero-Shot Learning.} The top panel shows the verbose initial prompt. The bottom panels compare the minimization trajectory of Qwen-32B (Left) and Llama-3.1-8B (Right) using Algorithm from Section \ref{sec:zsl}. Metrics indicate total score, BERT-score, and compression ratio. Running with Temperature 0.}
\label{fig:milestones_11334092_karim}
\end{figure*}
\begin{figure*}[t]
\centering
\begin{tcolorbox}[
    colback=mygray, 
    colframe=gray!50!black, 
    title=\textbf{Zero-Shot Learning: Initial Prompt}, 
    fonttitle=\sffamily\small, 
    sharp corners=south,
    boxrule=0.5pt,
    left=4pt, right=4pt, top=4pt, bottom=4pt
]
\small\sffamily\textit{ Explain the process of photosynthesis to 10th-grade biology students, including the major steps like light absorption by chlorophyll, the splitting of water molecules to release oxygen, and the Calvin cycle for sugar production. Make sure to cover how energy from the sun is converted into chemical energy stored in glucose, the role of stomata in gas exchange, and real-world examples of how disruptions like deforestation affect this process on a global scale, while keeping the language accessible and avoiding overly technical jargon.
}
\end{tcolorbox}
\vspace{-2pt}
\begin{minipage}[t]{0.49\textwidth}
\centering
\textcolor{myblue}{\textbf{\small Qwen/Qwen2.5-32B-Instruct-AWQ}}
\vspace{2pt}
\hrule height 0.8pt
\vspace{4pt}
\begin{flushleft}
\footnotesize
\textbf{Iter 1} {\scriptsize \color{gray} (Score: 0.2466 $\mid$ Bert: 0.9026 $\mid$ Comp: 0.3959)} \\ 
\vspace{1pt}
\texttt{Explain photosynthesis to 10th graders: light absorption, water splitting for oxygen, Calvin cycle for sugar. Show sun energy to glucose, stomata for gas exchange, and impact of deforestation. Use simple language.}\\
\vspace{4pt}
\textbf{Iter 2} {\scriptsize \color{gray} (Score: 0.2206 $\mid$ Bert: 0.9286 $\mid$ Comp: 0.3699)} \\ 
\vspace{1pt}
\texttt{Explain photosynthesis to 10th graders: light absorption, water splitting for oxygen, Calvin cycle for sugar. Show sun to glucose, stomata for gas exchange, deforestation impact. Use simple language.}\\
\vspace{4pt}
\textbf{Iter 3} {\scriptsize \color{gray} (Score: 0.2187 $\mid$ Bert: 0.8917 $\mid$ Comp: 0.3290)} \\ 
\vspace{1pt}
\texttt{Explain photosynthesis to 10th graders: light, water to oxygen, Calvin cycle for sugar. Show sun to glucose, stomata for gas exchange, deforestation impact. Use simple language.}\\
\vspace{4pt}
\textbf{Iter 4} {\scriptsize \color{gray} (Score: 0.2097 $\mid$ Bert: 0.8780 $\mid$ Comp: 0.2974)} \\ 
\vspace{1pt}
\texttt{Explain photosynthesis to 10th graders: light, water to oxygen, Calvin cycle for sugar. Show sun to glucose, stomata, deforestation impact. Use simple language.}\\
\vspace{4pt}
\textbf{Iter 5} {\scriptsize \color{gray} (Score: 0.1944 $\mid$ Bert: 0.9012 $\mid$ Comp: 0.2900)} \\ 
\vspace{1pt}
\texttt{Explain photosynthesis to 10th graders: light, water to oxygen, Calvin cycle for sugar. Show sun to glucose, stomata, deforestation impact. Simple language.}
\end{flushleft}
\end{minipage}
\hfill
\begin{minipage}[t]{0.49\textwidth}
\centering
\textcolor{myred}{\textbf{\small meta-llama/Llama-3.1-8B-Instruct}}
\vspace{2pt}
\hrule height 0.8pt
\vspace{4pt}
\begin{flushleft}
\footnotesize
\textbf{Iter 1} {\scriptsize \color{gray} (Score: 0.2027 $\mid$ Bert: 0.9310 $\mid$ Comp: 0.3364)} \\ 
\vspace{1pt}
\texttt{Explain photosynthesis to 10th-grade biology students, covering light absorption, water splitting, the Calvin cycle, energy conversion, stomata, and global impacts of deforestation.}\\
\vspace{4pt}
\textbf{Iter 2} {\scriptsize \color{gray} (Score: 0.1854 $\mid$ Bert: 0.9173 $\mid$ Comp: 0.2881)} \\ 
\vspace{1pt}
\texttt{Explain photosynthesis to 10th-grade biology students, covering light absorption, the Calvin cycle, energy conversion, and global impacts of deforestation.}\\
\vspace{4pt}
\textbf{Iter 3} {\scriptsize \color{gray} (Score: 0.1673 $\mid$ Bert: 0.9349 $\mid$ Comp: 0.2695)} \\ 
\vspace{1pt}
\texttt{Explain photosynthesis to 10th-grade biology students, covering light absorption, the Calvin cycle, energy conversion, and deforestation impacts.}\\
\vspace{4pt}
\textbf{Iter 4} {\scriptsize \color{gray} (Score: 0.1673 $\mid$ Bert: 0.9350 $\mid$ Comp: 0.2695)} \\ 
\vspace{1pt}
\texttt{Explain photosynthesis to 10th-grade biology students, covering light absorption, the Calvin cycle, energy conversion, and deforestation effects.}\\
\vspace{4pt}
\textbf{Iter 5} {\scriptsize \color{gray} (Score: 0.1496 $\mid$ Bert: 0.9425 $\mid$ Comp: 0.2416)} \\ 
\vspace{1pt}
\texttt{Explain photosynthesis to 10th graders, covering light absorption, the Calvin cycle, energy conversion, and deforestation impacts.}
\end{flushleft}
\end{minipage}
\vspace{6pt}
\caption{\textbf{Milestone Discoveries in Prompt Minimization using Zero-Shot Learning.} The top panel shows the verbose initial prompt. The bottom panels compare the minimization trajectory of Qwen-32B (Left) and Llama-3.1-8B (Right) using Algorithm from Section \ref{sec:zsl}. Metrics indicate total score, BERT-score, and compression ratio. Running with Temperature 0.}
\label{fig:milestones_4877789_karim}
\end{figure*}
\begin{figure*}[t]
\centering
\begin{tcolorbox}[
    colback=mygray, 
    colframe=gray!50!black, 
    title=\textbf{In-Context Learning: Initial Prompt}, 
    fonttitle=\sffamily\small, 
    sharp corners=south,
    boxrule=0.5pt,
    left=4pt, right=4pt, top=4pt, bottom=4pt
]
\small\sffamily\textit{ Explain the process of photosynthesis to 10th-grade biology students, including the major steps like light absorption by chlorophyll, the splitting of water molecules to release oxygen, and the Calvin cycle for sugar production. Make sure to cover how energy from the sun is converted into chemical energy stored in glucose, the role of stomata in gas exchange, and real-world examples of how disruptions like deforestation affect this process on a global scale, while keeping the language accessible and avoiding overly technical jargon.
}
\end{tcolorbox}
\vspace{-2pt}
\begin{minipage}[t]{0.49\textwidth}
\centering
\textcolor{myblue}{\textbf{\small Qwen/Qwen2.5-32B-Instruct-AWQ}}
\vspace{2pt}
\hrule height 0.8pt
\vspace{4pt}
\begin{flushleft}
\footnotesize
\textbf{Iter 1} {\scriptsize \color{gray} (Score: 0.2851 $\mid$ Bert: 0.9019 $\mid$ Comp: 0.4721)} \\ 
\vspace{1pt}
\texttt{Explain photosynthesis to 10th-grade students: light absorption by chlorophyll, water splitting to release oxygen, Calvin cycle for sugar. Cover energy conversion to glucose, role of stomata in gas exchange, impact of deforestation. Keep language simple.}\\
\vspace{4pt}
\textbf{Iter 2} {\scriptsize \color{gray} (Score: 0.2767 $\mid$ Bert: 0.8982 $\mid$ Comp: 0.4517)} \\ 
\vspace{1pt}
\texttt{Explain photosynthesis to 10th-grade students: light absorption by chlorophyll, water splitting to release oxygen, Calvin cycle for sugar. Cover energy conversion to glucose, stomata role, and impact of deforestation. Keep language accessible.}\\
\vspace{4pt}
\textbf{Iter 3} {\scriptsize \color{gray} (Score: 0.2743 $\mid$ Bert: 0.8919 $\mid$ Comp: 0.4405)} \\ 
\vspace{1pt}
\texttt{Explain photosynthesis to 10th-grade students: light absorption by chlorophyll, water splitting to release oxygen, Calvin cycle for sugar, energy conversion to glucose, stomata for gas exchange, impacts of deforestation. Keep accessible.}\\
\vspace{4pt}
\textbf{Iter 4} {\scriptsize \color{gray} (Score: 0.2724 $\mid$ Bert: 0.8958 $\mid$ Comp: 0.4405)} \\ 
\vspace{1pt}
\texttt{Explain photosynthesis to 10th-grade students: light absorption by chlorophyll, water splitting to release oxygen, Calvin cycle for sugar, energy conversion to glucose, stomata for gas exchange, impacts of deforestation. Keep accessible.}\\
\vspace{4pt}
\textbf{Iter 6} {\scriptsize \color{gray} (Score: 0.2719 $\mid$ Bert: 0.8967 $\mid$ Comp: 0.4405)} \\ 
\vspace{1pt}
\texttt{Explain photosynthesis to 10th-grade students: light absorption by chlorophyll, water splitting to release oxygen, Calvin cycle for sugar, energy conversion to glucose, stomata for gas exchange, impacts of deforestation. Keep accessible.}\\
\vspace{4pt}
\textbf{Iter 7} {\scriptsize \color{gray} (Score: 0.2713 $\mid$ Bert: 0.8978 $\mid$ Comp: 0.4405)} \\ 
\vspace{1pt}
\texttt{Explain photosynthesis to 10th-grade students: light absorption by chlorophyll, water splitting to release oxygen, Calvin cycle for sugar, energy conversion to glucose, stomata for gas exchange, impacts of deforestation. Keep accessible.}\\
\vspace{4pt}
\textbf{Iter 8} {\scriptsize \color{gray} (Score: 0.2705 $\mid$ Bert: 0.8995 $\mid$ Comp: 0.4405)} \\ 
\vspace{1pt}
\texttt{Explain photosynthesis to 10th-grade students: light absorption by chlorophyll, water splitting to release oxygen, Calvin cycle for sugar, energy conversion to glucose, stomata for gas exchange, impacts of deforestation. Keep accessible.}
\end{flushleft}
\end{minipage}
\hfill
\begin{minipage}[t]{0.49\textwidth}
\centering
\textcolor{myred}{\textbf{\small meta-llama/Llama-3.1-8B-Instruct}}
\vspace{2pt}
\hrule height 0.8pt
\vspace{4pt}
\begin{flushleft}
\footnotesize
\textbf{Iter 1} {\scriptsize \color{gray} (Score: 0.1817 $\mid$ Bert: 0.8819 $\mid$ Comp: 0.2454)} \\ 
\vspace{1pt}
\texttt{**Step-by-step explanation of photosynthesis for 10th-grade biology students: 5 steps, accessible language, real-world disruptions**}\\
\vspace{4pt}
\textbf{Iter 2} {\scriptsize \color{gray} (Score: 0.1532 $\mid$ Bert: 0.8795 $\mid$ Comp: 0.1859)} \\ 
\vspace{1pt}
\texttt{**Step-by-step explanation of photosynthesis: 5 steps, accessible language, real-world disruptions**}\\
\vspace{4pt}
\textbf{Iter 3} {\scriptsize \color{gray} (Score: 0.1403 $\mid$ Bert: 0.8717 $\mid$ Comp: 0.1524)} \\ 
\vspace{1pt}
\texttt{**5-step photosynthesis explanation: accessible language, real-world disruptions**}\\
\vspace{4pt}
\textbf{Iter 4} {\scriptsize \color{gray} (Score: 0.1286 $\mid$ Bert: 0.8711 $\mid$ Comp: 0.1283)} \\ 
\vspace{1pt}
\texttt{5-step photosynthesis explanation, 10th-grade, real-world disruptions}\\
\vspace{4pt}
\textbf{Iter 5} {\scriptsize \color{gray} (Score: 0.1212 $\mid$ Bert: 0.8766 $\mid$ Comp: 0.1190)} \\ 
\vspace{1pt}
\texttt{**5-step explanation of photosynthesis, real-world disruptions**}\\
\vspace{4pt}
\textbf{Iter 6} {\scriptsize \color{gray} (Score: 0.1089 $\mid$ Bert: 0.8733 $\mid$ Comp: 0.0911)} \\ 
\vspace{1pt}
\texttt{Photosynthesis in 5 steps, real-world disruptions}\\
\vspace{4pt}
\textbf{Iter 8} {\scriptsize \color{gray} (Score: 0.1036 $\mid$ Bert: 0.8765 $\mid$ Comp: 0.0836)} \\ 
\vspace{1pt}
\texttt{**Photosynthesis in 5 steps and disruptions**}\\
\vspace{4pt}
\textbf{Iter 9} {\scriptsize \color{gray} (Score: 0.1016 $\mid$ Bert: 0.8804 $\mid$ Comp: 0.0836)} \\ 
\vspace{1pt}
\texttt{**Photosynthesis in 5 steps and disruptions**}\\
\vspace{4pt}
\textbf{Iter 10} {\scriptsize \color{gray} (Score: 0.0977 $\mid$ Bert: 0.8882 $\mid$ Comp: 0.0836)} \\ 
\vspace{1pt}
\texttt{**Photosynthesis in 5 steps and disruptions**}\\
\vspace{4pt}
\textbf{Iter 12} {\scriptsize \color{gray} (Score: 0.0907 $\mid$ Bert: 0.8818 $\mid$ Comp: 0.0632)} \\ 
\vspace{1pt}
\texttt{Photosynthesis 5 steps, accessible}\\
\vspace{4pt}
\textbf{Iter 13} {\scriptsize \color{gray} (Score: 0.0841 $\mid$ Bert: 0.8801 $\mid$ Comp: 0.0483)} \\ 
\vspace{1pt}
\texttt{Photosynthesis in 5 steps.}\\
\vspace{4pt}
\textbf{Iter 14} {\scriptsize \color{gray} (Score: 0.0786 $\mid$ Bert: 0.8893 $\mid$ Comp: 0.0465)} \\ 
\vspace{1pt}
\texttt{Photosynthesis in 5 steps}
\end{flushleft}
\end{minipage}
\vspace{6pt}
\caption{\textbf{Milestone Discoveries in Prompt Minimization using In-Context Learning.} The top panel shows the verbose initial prompt. The bottom panels compare the minimization trajectory of Qwen-32B (Left) and Llama-3.1-8B (Right) using Algorithm from Section \ref{sec:icl}. Metrics indicate total score, BERT-score, and compression ratio. Running with Temperature 0.9.}
\label{fig:milestones_4877789_marius}
\end{figure*}
\begin{figure*}[t]
\centering
\begin{tcolorbox}[
    colback=mygray, 
    colframe=gray!50!black, 
    title=\textbf{RL Fine-Tuning: Initial Prompt}, 
    fonttitle=\sffamily\small, 
    sharp corners=south,
    boxrule=0.5pt,
    left=4pt, right=4pt, top=4pt, bottom=4pt
]
\small\sffamily\textit{ As artificial intelligence systems continue integrating into nearly every aspect of daily life—from personalized assistants that anticipate our needs to automated systems that influence hiring, finance, healthcare, and public policy—the question of how humans and machines should coexist becomes increasingly complex. Beyond simply determining when machines should take over tasks, society must grapple with how AI reshapes human agency, autonomy, and social structures. Considering the tradeoffs between convenience, efficiency, and control, how do you envision the ideal balance between human judgment and machine autonomy? What cultural shifts, safeguards, regulatory frameworks, or ethical principles do you believe are necessary to ensure that these technologies enhance human well-being while protecting individual freedoms, preventing algorithmic biases, and preserving meaningful human oversight?
}
\end{tcolorbox}
\vspace{-2pt}
\begin{minipage}[t]{0.49\textwidth}
\centering
\textcolor{myblue}{\textbf{\small Qwen/Qwen2.5-32B-Instruct-AWQ}}
\vspace{2pt}
\hrule height 0.8pt
\vspace{4pt}
\begin{flushleft}
\footnotesize
\textbf{Iter 1} {\scriptsize \color{gray} (Score: 0.2293 $\mid$ Bert: 0.8744 $\mid$ Comp: 0.3330)} \\ 
\vspace{1pt}
\texttt{Analyze AI's impact on human life in areas like assistants, hiring, finance, healthcare, \& policy. Evaluate balance btwn human judgment \& machine autonomy. Identify nec. cultural shifts, safeguards, regulations, \& ethics to enhance well-being, protect freedoms, prevent bias, \& ensure human oversight.}\\
\vspace{4pt}
\textbf{Iter 2} {\scriptsize \color{gray} (Score: 0.2257 $\mid$ Bert: 0.8926 $\mid$ Comp: 0.3440)} \\ 
\vspace{1pt}
\texttt{AI integrates into daily life (assistants, hiring, finance, healthcare, policy). Define ideal balance btwn human judgment \& machine autonomy. Consider conven, effi, ctrl. Suggest cultural shifts, safeguards, reg frameworks, ethics to enhance well-being, protect freedoms, prevent biases, ensure human oversight.}\\
\vspace{4pt}
\textbf{Iter 3} {\scriptsize \color{gray} (Score: 0.2217 $\mid$ Bert: 0.8620 $\mid$ Comp: 0.3053)} \\ 
\vspace{1pt}
\texttt{Analyze AI integration impact on human life, task delegation, agency, \& social struct. Discuss balance btw convenience, efficiency, control. Suggest nec. cultural shifts, safeguards, regs, ethics to enhance well-being, protect freedoms, prevent biases, ensure human oversight.}\\
\vspace{4pt}
\textbf{Iter 5} {\scriptsize \color{gray} (Score: 0.2092 $\mid$ Bert: 0.8891 $\mid$ Comp: 0.3075)} \\ 
\vspace{1pt}
\texttt{AI integrates into daily life (assistants, hiring, finance, healthcare, policy). Discuss balance between human judgment \& machine autonomy. Outline cultural shifts, safeguards, regulations, ethics needed to enhance well-being, protect freedoms, prevent biases, ensure oversight.}\\
\vspace{4pt}
\textbf{Iter 13} {\scriptsize \color{gray} (Score: 0.1979 $\mid$ Bert: 0.8984 $\mid$ Comp: 0.2942)} \\ 
\vspace{1pt}
\texttt{AI integrates into daily life (assistants, hiring, finance, healthcare, policy). Define balance btwn human judgment \& machine autonomy. Identify nec. cultural shifts, safeguards, regs, ethics to enhance well-being, protect freedoms, prevent biases, ensure oversight.}\\
\end{flushleft}
\end{minipage}
\hfill
\begin{minipage}[t]{0.49\textwidth}
\centering
\textcolor{myred}{\textbf{\small meta-llama/Llama-3.1-8B-Instruct}}
\vspace{2pt}
\hrule height 0.8pt
\vspace{4pt}
\begin{flushleft}
\footnotesize
\textbf{Iter 1} {\scriptsize \color{gray} (Score: 0.1178 $\mid$ Bert: 0.8806 $\mid$ Comp: 0.1162)} \\ 
\vspace{1pt}
\texttt{\textgreater{} Determine the ideal balance between human judgment \& machine autonomy as AI integrates into daily life.}\\
\vspace{4pt}
\textbf{Iter 7} {\scriptsize \color{gray} (Score: 0.1166 $\mid$ Bert: 0.8852 $\mid$ Comp: 0.1184)} \\ 
\vspace{1pt}
\texttt{Determine the ideal balance between human judgment \& machine autonomy in an increasingly AI-driven society.}\\
\vspace{4pt}
\textbf{Iter 10} {\scriptsize \color{gray} (Score: 0.1110 $\mid$ Bert: 0.8876 $\mid$ Comp: 0.1095)} \\ 
\vspace{1pt}
\texttt{\textgreater{} Determine the ideal balance between human judgment \& machine autonomy in an AI-dominated society.}\\
\vspace{4pt}
\textbf{Iter 11} {\scriptsize \color{gray} (Score: 0.1014 $\mid$ Bert: 0.8746 $\mid$ Comp: 0.0774)} \\ 
\vspace{1pt}
\texttt{Determine the ideal balance between human judgment \& machine autonomy.}\\
\vspace{4pt}
\textbf{Iter 14} {\scriptsize \color{gray} (Score: 0.0973 $\mid$ Bert: 0.8783 $\mid$ Comp: 0.0730)} \\ 
\vspace{1pt}
\texttt{\textgreater{} Determine the balance between human judgment \& machine autonomy.}\\
\end{flushleft}
\end{minipage}
\vspace{6pt}
\caption{\textbf{Milestone Discoveries in Prompt Minimization using RL Fine-Tuning.} The top panel shows the verbose initial prompt. The bottom panels compare the minimization trajectory of Qwen-32B (Left) and Llama-3.1-8B (Right) using Algorithm from Section \ref{sec:rl}. Metrics indicate total score, BERT-score, and compression ratio. Running with Temperature 0.8.}
\label{fig:milestones_5129854_li}
\end{figure*}
\begin{figure*}[t]
\centering
\begin{tcolorbox}[
    colback=mygray, 
    colframe=gray!50!black, 
    title=\textbf{RL Fine-Tuning: Initial Prompt}, 
    fonttitle=\sffamily\small, 
    sharp corners=south,
    boxrule=0.5pt,
    left=4pt, right=4pt, top=4pt, bottom=4pt
]
\small\sffamily\textit{ Explain the process of photosynthesis to 10th-grade biology students, including the major steps like light absorption by chlorophyll, the splitting of water molecules to release oxygen, and the Calvin cycle for sugar production. Make sure to cover how energy from the sun is converted into chemical energy stored in glucose, the role of stomata in gas exchange, and real-world examples of how disruptions like deforestation affect this process on a global scale, while keeping the language accessible and avoiding overly technical jargon.
}
\end{tcolorbox}
\vspace{-2pt}
\begin{minipage}[t]{0.49\textwidth}
\centering
\textcolor{myblue}{\textbf{\small Qwen/Qwen2.5-32B-Instruct-AWQ}}
\vspace{2pt}
\hrule height 0.8pt
\vspace{4pt}
\begin{flushleft}
\footnotesize
\textbf{Iter 1} {\scriptsize \color{gray} (Score: 0.2161 $\mid$ Bert: 0.8931 $\mid$ Comp: 0.3253)} \\ 
\vspace{1pt}
\texttt{Explain photosynthesis to 10th graders: light吸收by叶绿素,水分子split to release O₂, Calvin cycle for糖production. Cover阳光转化学能in葡萄糖,气孔role in气体交换, \& how破坏如伐木影响全球光合作用. Use accessible语言.}\\
\end{flushleft}
\end{minipage}
\hfill
\begin{minipage}[t]{0.49\textwidth}
\centering
\textcolor{myred}{\textbf{\small meta-llama/Llama-3.1-8B-Instruct}}
\vspace{2pt}
\hrule height 0.8pt
\vspace{4pt}
\begin{flushleft}
\footnotesize
\textbf{Iter 1} {\scriptsize \color{gray} (Score: 0.2750 $\mid$ Bert: 0.9165 $\mid$ Comp: 0.4665)} \\ 
\vspace{1pt}
\texttt{Explain photosynthesis to 10th-grade biology students, covering light absorption by chlorophyll, water molecule splitting, oxygen release, Calvin cycle for sugar production, energy conversion, stomata function, \& global disruptions like deforestation.}\\
\vspace{4pt}
\textbf{Iter 3} {\scriptsize \color{gray} (Score: 0.2663 $\mid$ Bert: 0.9488 $\mid$ Comp: 0.4814)} \\ 
\vspace{1pt}
\texttt{Explain photosynthesis to 10th-grade biology students, covering light absorption by chlorophyll, water molecule splitting \& oxygen release, the Calvin cycle, energy conversion to glucose, stomata's role in gas exchange \& global disruptions like deforestation.}\\
\end{flushleft}
\end{minipage}
\vspace{6pt}
\caption{\textbf{Milestone Discoveries in Prompt Minimization using RL Fine-Tuning.} The top panel shows the verbose initial prompt. The bottom panels compare the minimization trajectory of Qwen-32B (Left) and Llama-3.1-8B (Right) using Algorithm from Section \ref{sec:rl}. Metrics indicate total score, BERT-score, and compression ratio. Running with Temperature 0.8.}
\label{fig:milestones_59082115_li}
\end{figure*}

\end{document}